\documentclass{article} 
\usepackage{iclr2021_conference,times}

\usepackage[utf8]{inputenc}                                 
\usepackage[T1]{fontenc}                                    
\usepackage{xcolor}         
\definecolor{custom-blue}{RGB}{6,69,173} 
\usepackage[colorlinks, allcolors=custom-blue]{hyperref}
\usepackage{url}                                            
\usepackage{booktabs}                                       
\usepackage{amsfonts}                                       
\usepackage{amssymb}                                        
\usepackage{amsthm}    									  
\usepackage{amsmath}                                     
\usepackage{mathtools}                                      
\usepackage{nicefrac}                                       
\usepackage{microtype}                                      
\usepackage{bm}                                             
\usepackage{lmodern}                                        %
\usepackage[font=small,labelfont=bf,textfont=it]{caption}   
\usepackage{subcaption}                                     
\usepackage[ruled,vlined]{algorithm2e}                      
\usepackage{float}
\usepackage{multirow}
\usepackage{xcolor,pifont}
\newcommand*\colourcheck[1]{%
  \expandafter\newcommand\csname #1check\endcsname{\textcolor{#1}{\ding{52}}}%
}
\colourcheck{green}

\newcommand*\colourx[1]{%
  \expandafter\newcommand\csname #1x\endcsname{\textcolor{#1}{\ding{55}}}%
}
\colourx{red}

\setcitestyle{authoryear,open={(},close={)}}

\theoremstyle{plain}
\newtheorem{definition}{Theorem}

\title{
Adversarial Boot Camp: label free certified robustness in one epoch}

\usepackage{authblk}
\makeatletter
\renewcommand\AB@authnote[1]{\rlap{\textsuperscript{\normalfont#1}}}

\makeatother 

\author[1*]{\textbf{Ryan Campbell}\thanks{\textsuperscript{*} Correspondance to
\texttt{ryan.campbell2@mail.mcgill.ca}}}
\author[]{\textbf{Chris Finlay}}
\author[]{\textbf{Adam M Oberman}}
\affil[]{Department of Mathematics and Statistics, McGill University,
Montr\'eal, Canada}
\renewcommand\footnotemark{}

\iclrfinalcopy 
\begin{document}

\maketitle

\begin{abstract}
Machine learning models are vulnerable to adversarial attacks.  One approach to addressing this vulnerability is certification, which focuses on models that are guaranteed to be robust for a given perturbation size.  A drawback of recent certified models is that they are stochastic: they require multiple computationally expensive model evaluations with random noise added to a given input. In our work, we present a \emph{deterministic} certification approach which results in a certifiably robust model. This approach is based on an equivalence between training with a particular regularized loss, and the expected values of Gaussian averages. We achieve certified models on ImageNet-1k by retraining a model with this loss for one epoch without the use of label information.

\end{abstract}

%
%

\section{Introduction}
Neural networks are very accurate on image classification tasks, but they are
vulnerable to adversarial perturbations, i.e. small changes to the model input leading to misclassification~\citep{szegedy2013intriguing}. 
Adversarial training \citep{madry2017towards} improves robustness, at the expense of a loss of accuracy on unperturbed images \citep{zhang2019theoretically}. Model certification \citep{lecuyer2019certified, raghunathan2018certified, cohen2019certified} is complementary approach to adversarial training,  which provides a guarantee that a model prediction is invariant to perturbations up to a given norm. 

Given an input $x$, the model $f$ is certified to $\ell_2$ norm $r$ at $x$ if 
it gives the same classification on $f(x+\eta)$ for all perturbation $\eta$ with norm up to $r$, 
\begin{equation}\label{cert_intro}
\arg\max f(x + \eta) = \arg\max f(x), \quad \text{ for all } \|\eta\|_2 \leq r	
\end{equation}
\cite{cohen2019certified} and \cite{salman2019provably} certify models by defining a ``smoothed'' model, $f^{smooth}$, which is the expected Gaussian average of our initial model $f$ at a given input example $x$,
\begin{equation}
	\label{eq:smoothed}
	f^{smooth}(x) \approx \mathbb{E}_\eta \left[f(x+\eta)\right]
\end{equation}
where the perturbation is sampled from a Gaussian, $\eta\sim\mathcal{N}(0,\sigma^2I)$.
\cite{cohen2019certified} used a probabilistic argument to show that models defined by \eqref{eq:smoothed} can be certified to a given radius by making a large number of stochastic model evaluations. Certified models can classify by first averaging the model, \citep{salman2019provably}, or by taking the mode, the most popular classification given by the ensemble \citep{cohen2019certified}.


\citeauthor{cohen2019certified} and \citeauthor{salman2019provably} approximate the model $f^{smooth} $ stochastically, using a Gaussian ensemble, which consists of evaluating the base model $f$ multiple times on the image perturbed by noise.  Like all ensemble models, these stochastic models require multiple inferences, which is more costly than performing inference a single time.  In addition,  these stochastic models require training the base model $f$ from scratch, by exposing it to Gaussian noise, in order to improve the accuracy of $f^{smooth}$.    \cite{salman2019provably} additionally expose the model to adversarial attacks during training.  In the case of certified models, there is a trade-off between certification and accuracy: the certified models lose accuracy on unperturbed images. 

\begin{figure}[t]
    \centering
    \begin{subfigure}{.49\textwidth}
    \centering
    		\includegraphics[width=\linewidth]{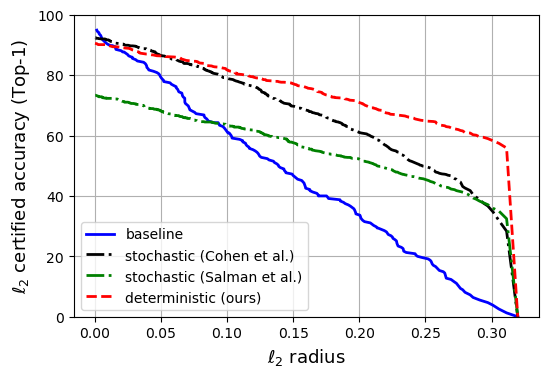}
    		\caption{CIFAR-10 top-1 certified accuracy, $\sigma=0.10$}
    		\label{fig:cifar-cert}
	\end{subfigure}
    \begin{subfigure}{.49\textwidth}
    \centering
    		\includegraphics[width=\linewidth]{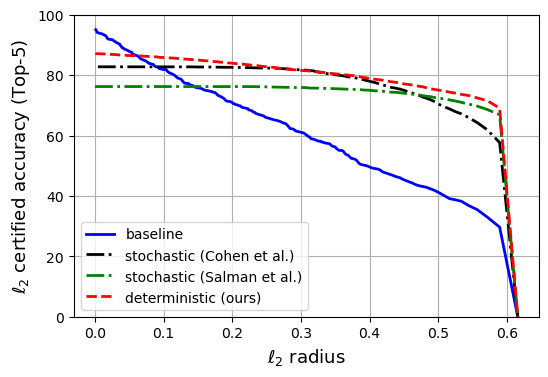}
    		\caption{ImageNet top-5 certified accuracy, $\sigma=0.25$}
    		\label{fig:imagenet-cert}
    \end{subfigure}
    \caption{Certified accuracy as a function of $\ell_2$ radius.}
    \label{fig:cert}
\end{figure}

In this work, we present a \emph{deterministic} model $f^{smooth}(x)$ given by \eqref{eq:smoothed}.  Unlike the stochastic models, we do not need to train a new model from scratch with Gaussian noise data augmentations; instead, we can fine tune an accurate baseline model for one epoch, using a loss designed to encourage \eqref{eq:smoothed} to hold.  The result is a certified deterministic model which is \emph{faster to train, and faster to perform inference}. In addition, the certification radius for our model is improved, compared to previous work, on both the CIFAR-10 and ImageNet databases, see~\autoref{fig:cert}.  Moreover, the accuracy on unperturbed images is improved on the ImageNet dataset \citep{deng2009imagenet} with a Top-5 accuracy of 86.1\% for our deterministic model versus  85.3\% for \cite{cohen2019certified} and 78.8\% for \cite{salman2019provably}. 

To our knowledge, this is the first deterministic Gaussian smoothing certification
technique. The main appeal of our approach is a large  decrease in the number of function
evaluations for computing certifiably robust models, and a
corresponding decrease in compute time at inference. Rather than stochastically sampling many times
from a Gaussian at inference time, our method is certifiably robust with only a
single model query. This greater speed in model evaluation is demonstrated in Table~\ref{tab:comp-time}.  Moreover, the deterministic certified model can be obtained by retraining a model for one epoch, without using labels. 
The speed and flexibility of this method allows it to be used to make empirically robust models certifiably robust. To see this, we test our method on adversarially trained models from \cite{madry2017towards}, increasing the certified radius of the adversarially robust model, see \autoref{tab:cert-madry}.



\begin{table}[h!]
  \caption{A comparison of robust models.  Stochastic smoothing arises from methods like the ones presented in \cite{cohen2019certified} and \cite{salman2019provably}. Adversarial training from \cite{madry2017towards}.}
  \label{tab:compare-methods}
  \centering
  \begin{tabular}{lccc}
    \toprule
    \multirow{2}{*}{Model}     & Can be obtained from & Evaluation in one & \multirow{2}{*}{Is certified?} \\
    & any pretrained model  & forward pass & \\
    \midrule
    Deterministic Smoothing (ours)  &  \greencheck  &  \greencheck  &  \greencheck   \\
    Stochastic Smoothing  & \redx   &  \redx  &  \greencheck   \\
    Adversarial Training & \redx & \greencheck &  \redx \\
    \bottomrule
  \end{tabular}
\end{table}

\begin{table}[h!]
  \caption{Average classification inference time (seconds)}
  \label{tab:comp-time}
  \centering
  \begin{tabular}{lcccc}
    \toprule
    Model     &  \multicolumn{2}{c}{CIFAR-10}  & \multicolumn{2}{c}{ImageNet-1k}  \\
     & CPU & GPU & CPU & GPU\\
    \midrule
    Deterministic (ours)  &  0.0049 & 0.0080  &  0.0615 & 0.0113  \\
    Stochastic \citep{cohen2019certified}  & 0.0480& 0.0399 &  0.1631  & 0.0932\\
    \bottomrule
  \end{tabular}
\end{table}

\section{Related work}


The issue of adversarial vulnerability arose in the works of
\citet{szegedy2013intriguing} and \citet{goodfellow2014explaining}, and has
spawned a vast body of research. The idea of training models to be robust to adversarial attacks was widely popularized in \cite{madry2017towards}. This method, known as \emph{adversarial training}, trains a model on images corrupted by gradient-based adversarial attacks, resulting in robust models.
In terms of certification, early work by
\citet{cheng2017maximum} provided a method of
computing maximum perturbation bounds for neural networks, and reduced to
solving a mixed integer optimization problem. \cite{weng2018certification}
introduced non-trivial robustness bounds for fully connected networks, and provided tight robustness bounds at low computational cost. 
\cite{weng2018clever} proposed a metric that has theoretical grounding based on
Lipschitz continuity of the classifier model and is scaleable to state-of-the-art
ImageNet neural network classifiers. \cite{zhang2018crown} proposed a
general framework to certify neural networks based on linear and quadratic
bounding techniques on the activation functions, which is
more flexible than its predecessors.

    
Training a neural network with Gaussian noise has been shown to be equivalent to
gradient regularization \citep{bishop1995training}. This helps improve
robustness of models; however, recent work has used additive noise during training and evaluation for certification purposes. \cite{lecuyer2019certified} first considered adding
random Gaussian noise as a certifiable defense in a method called
\textit{PixelDP}. In their method, they take a known neural network architecture
and add a layer of random noise to make the model's output random. The expected
classification is in turn more robust to adversarial perturbations. Furthermore,
their defense is a certified defense, meaning they provide a lower bound on the
amount of adversarial perturbations for which their defence will always work.
In a following work, \cite{li2018second} provided a defence with improved certified robustness. 
The certification guarantees given in these two papers are loose, meaning the
defended model will always be more robust than the certification bound indicates.

In contrast, \cite{cohen2019certified} provided a defence utilizing randomized Gaussian
smoothing that leads to \emph{tight} robustness guarantees under the $\ell_2$
norm. Moreover \citeauthor{cohen2019certified} used
Monte Carlo sampling to compute the radius in which a model's prediction is
unchanged; we refer to this method as \textsc{RandomizedSmoothing}.  
In work building on \citeauthor{cohen2019certified}, \cite{salman2019provably}
developed an adversarial training
framework called \textsc{SmoothAdv}  and defined a Lipschitz constant of averaged models. \cite{yang2020randomized} generalize previous randomized smoothing methods by providing robustness guarantees in the $\ell_1$, $\ell_2$, and $\ell_\infty$ norms for smoothing with several non-Gaussian distributions.


\section{Deterministic Smoothing}
Suppose we are given a dataset consisting of paired samples
$(x,y)\in\mathcal{X}\times\mathcal{Y}$ where $x$ is an example with corresponding true classification $y$. The supervised learning approach trains
a model $f:\mathcal{X}\longrightarrow\mathbb{R}^{Nc}$ which maps images to a
vector whose length equals the number of classes. Suppose  $f$  is the initial model,
and let $f^{smooth}$ be the averaged model given by Equation \eqref{eq:smoothed}. \citet{cohen2019certified} find a Gaussian smoothed classification model $f^{smooth}$  by sampling $\eta\sim\mathcal{N}(0,\sigma^2I)$ independently $n$ times, performing $n$ classifications, and then computing the most popular classification. In the randomized smoothing method, the initial model $f$ is trained on data which is augmented with Gaussian noise to improve accuracy on noisy
images.

We take a different approach to Gaussian smoothing.
Starting from an accurate pretrained model $f$, we now discard the training
labels, and iteratively retrain a new model, $f^{smooth}$ using a quadratic loss
between the model $f$ and the new model's predictions, with an
additional gradient regularization term.   
We have found that discarding the original one-hot
labels and instead using model predictions helps make the model smoother. 

To be precise, our new models is a result of minimizing the loss which we call \textsc{HeatSmoothing},
\begin{equation}
    \label{eq:final-variational}
  \mathbb{E}_x\left[ \frac{1}{2}\left\|\text{softmax}\left(f^{smooth}(x)\right)-\text{softmax}\left(f(x)\right)\right\|_2^2 + \frac{\sigma^2}{2}\left\|\nabla_x f^{smooth}(x)\right\|_2^2\right]
\end{equation}
The smoothing achieved by the new models is illustrated schematically in Figure~\ref{fig:heat-eqn-solv}. 

\subsection{Related regularized losses}
Gradient regularization is known to be equivalent to Gaussian smoothing  \citep{bishop1995training, lecun1998LeNetpaper}.
Our deterministic smoothed
  model arises by training using the \textsc{HeatSmoothing} loss \eqref{eq:final-variational},
which is designed so to ensure that \eqref{eq:smoothed} holds for our model. 
Our results is related to the early results on regularized networks \citep{bishop1995training, lecun1998LeNetpaper}: that full gradient regularization is equivalent to Gaussian smoothing.  Formally this is stated as follows.
\begin{definition}\label{thm:Bishop}
\citep{bishop1995training} Training a feed-forward neural-network model using the quadratic (or mean-squared error) loss, with added Gaussian noise of mean 0 and variance $\sigma^2$ to the inputs, is equivalent to training with 
\begin{equation}
	\label{Noise_Tychonoff}
	\mathbb E _x\left[\|f(x) - y\|^2
  + \sigma^2 \| \nabla f(x)\|^2  \right ] 
\end{equation}
up to higher order terms. 
\end{definition}
The equivalence is normally used to go from models augmented with Gaussian noise to regularized models. Here, we use the result in the other direction; we train a regularized model in order to produce a model which is equivalent to evaluating with noise. In practice, this means that rather than adding noise to regularize models for
certifiable robustness, we explicitly
perform a type of gradient regularization, \emph{in order to produce a model
which  performs as if Gaussian noise was added.} See \autoref{fig:Cartoon} in \autoref{app:train-plot} for an illustration of the effect of this gradient regularization. 

The gradient regularization term in the \textsc{HeatSmoothing} loss \eqref{eq:final-variational}, is also related to adversarial training.   Tikhonov regularization is used to produced adversarially robust models \citep{finlay2019scaleable}. 
However in adversarial training, the gradient of the loss is used, rather that the gradient of the full model. Also,  our loss does not use information from the true labels. The reason for these differences is due to the fact that we wish to have a model that approximates the Gaussian average of our initial model $f$ (see Appendix~\ref{app:proof-heat-thm}). Furthermore, minimizing the gradient-norm of the loss of the output gives us a smooth model in all directions, rather than being robust to only adversarial perturbations.
%
%

\subsection{Algorithmic Details}
We have found that early on in training, the  value
$\frac{1}{2}\left\|f^{smooth}(x)-f^k(x)\right\|_2^2$ may be far greater than the
$\frac{\sigma^2}{2}\left\|\nabla_x f^{smooth}(x)\right\|_2^2$ term. So we introduced a softmax of the vectors in the distance-squared term to reduce the overall magnitude of this term.
We perform the training minimization of \eqref{eq:final-variational} for one epoch. 
The pseudo-code for our neural network weight update is given by Algorithm \ref{alg:heat-smoothing} 
\footnote{Code and links to trained models are publicly available at \url{https://github.com/ryancampbell514/HeatSmoothing/}}
%

Note that the $\left\|\nabla_x f^{smooth}(x)\right\|_2^2$ term in
\eqref{eq:final-variational} requires the computation of a Jacobian matrix norm. In
high dimensions this is computationally expensive. To approximate this term, we
make use of the \textit{Johnson-Lindenstrauss lemma}
\citep{johnson1984extensions, vempala2005johnson-lindenstrauss} followed by the
finite difference approximation from \cite{finlay2019scaleable}. We are able to
approximate $\left\|\nabla_x f^{smooth}(x)\right\|_2^2$ by taking the average of the
product of the Jacobian matrix and Gaussian noise vectors. Jacobian-vector
products can be easily computed via reverse mode automatic differentiation, by
moving the noise vector $w$ inside: 
\begin{equation}
	\label{eq:moving-noise}
	w\cdot\left(\nabla_x v(x)\right)= \nabla_x \left(w\cdot v(x)\right)
\end{equation}
Further computation expense is reduced by using finite-differences to
approximate the norm of the gradient. Once the finite-difference is computed, we
detach this term from the automatic differentiation computation graph, further speeding training. More details of our implementation of these approximation techniques, and the definition of the term $\hat{g}$ which is a regularization of the gradient,  are presented in Appendix~\ref{app:jl}.

\subsection{Theoretical details}

\begin{algorithm}[t!]
 	\caption{\textsc{HeatSmoothing} Neural Network Weight Update}
    \label{alg:heat-smoothing}
    \SetKwInOut{Input}{Input}
    \SetKwInOut{Update}{Update}
    \SetKwInOut{Compute}{Compute}
    \SetKwInOut{Output}{Output}
	\SetKwInOut{Initialize}{Initialize}
	\SetKwInOut{Return}{Return}
	\SetKw{Continue}{continue}
	\SetKw{Break}{break}
    \SetAlgoLined
    \Input{Minibatch of input examples $\bm{x}^{(mb)}=\left(x^{(1)},\dots,x^{(Nb)}\right)$\\ 
    A model $v$ set to ``train'' mode \\ 
    Current model $f$ set to ``eval'' mode \\ 
    $\sigma$, standard deviation of Gaussian smoothing \\
    $\kappa$, number of Gaussian noise replications (default$=10$) \\
    $\delta$, finite difference step-size (default$=0.1$)}
    \Update{learning-rate according to a pre-defined scheduler.}
    \For{$i\in\left\{1,\dots Nb\right\}$}{
    \Compute{$f^{smooth}(x^{(i)}),f(x^{(i)})\in\mathbb{R}^{Nc}$\\
    		$J_i = \frac{1}{2}\left\|f^{smooth}(x^{(i)})-f(x^{(i)})\right\|_2^2\in\mathbb{R}$\\
    		\For{$j\in\left\{1,\dots\kappa\right\}$}{
    			Generate $w = \frac{1}{\sqrt{Nc}}\left(w_1,\dots,w_{Nc}\right)$, $w_1,\dots,w_{Nc}\in\mathcal{N}(0,1)$\\
    			Compute the normalized gradient $\hat{g}$ via \eqref{eq:hatg}\\
    			Detach $x^{(i)}$ from the computation graph\\
			$J_i \leftarrow J_i + \dfrac{\sigma^2}{2 \delta^2}\left({w\cdot f^{smooth}(x^{(i)}+\delta \hat{g}) - w\cdot f^{smooth}(x^{(i)})}\right)^2$    		
    		}
    		}
    		$J\leftarrow \frac{1}{Nb}\sum\limits_{i=1}^{Nb} J_i$}
    Update the weights of $v$ by running backpropagation on $J$ at the current learning rate.
\end{algorithm}
%

We appeal to partial differential equations (PDE) theory for explaining the
equivalence between gradient regularization and Gaussian convolution (averaging) of the
model\footnote{We sometimes interchange the terms Gaussian averaging and Gaussian
convolution; they are equivalent, as shown in Theorem
\ref{thm:heat-equiv}.}. The idea is that the gradient term which appears in the loss leads to a smoothing of the new function (model).  The fact that the exact form of the
smoothing corresponds to Gaussian convolution is a mathematical result which can be interpreted probabilistically or using techniques from analysis. Briefly, we detail the link as follows.

\cite{einstein1906theory} showed that the function value of an averaged model
under Brownian motion is related to
the heat equation (a PDE); the theory of stochastic differential equations makes this
rigorous \citep{karatzas1998brownian}. Moreover, solutions of the heat
equation are given by Gaussian convolution with the original model.
Crucially, solutions of the
heat equation can be interpreted as iterations of a regularized loss problem
(called a variational energy) like that of Equation \eqref{eq:final-variational}.  The
  minimizer of this variational energy
\eqref{eq:final-variational} satisfies an equation which is formally equivalent
to the heat equation \citep{gelfand2000calculus}. Thus, taking these facts
together, we see that a few steps of the minimization of the loss in
\eqref{eq:final-variational} yields a model which approximately satisfies the
heat equation, and corresponds to a model smoothed by Gaussian convolution. See
\autoref{fig:Cartoon} for an illustration of a few steps of the training
procedure. This result is summarized in the following theorem.
\begin{definition}\label{thm:heat-equiv} \citep{strauss2007partial} Let $f$ be a bounded function, $x\in\mathbb{R}^d$, and $\eta\sim\mathcal{N}\left(0,\sigma^2I\right)$. Then the following are equivalent:
    \begin{enumerate}
        \item $\mathbb{E}_\eta\left[f(x+\eta)\right]$, the expected value of Gaussian averages of $f$ at $x$.
        \item $\left(f\ast\mathcal{N}(0,\sigma^2I)\right)(x)$, the convolution of $f$ with the density of the $\mathcal{N}(0,\sigma^2I)$ distribution evaluated at $x$.
        \item The  solution of the heat equation,
        \begin{equation}
            \label{eq:heat-eq}
            \frac{\partial}{\partial t}f(x,t) = \frac{\sigma^2}{2}\Delta_x f(x,t)
        \end{equation}
        at time $t=1$, with initial condition $f(x,0)=f(x)$.
    \end{enumerate}
\end{definition}

	 
In Appendix~\ref{app:proof-heat-thm}, we use Theorem~\ref{thm:heat-equiv} to show the equivalence of training with noise and iteratively training \eqref{eq:final-variational}.

To assess how well our model approximates the Gaussian average of the initial
model, we compute the certified $\ell_2$ radius for averaged models introduced in
\cite{cohen2019certified}. A larger radius implies a better approximation of the
Gaussian average of the initial model. We compare our models with stochastically
averaged models via \textit{certified accuracy}. This is the fraction of the test set
which a model correctly classifies at a given radius while ignoring abstained
classifications. Throughout, we always use the same $\sigma$ value for certification as for
training. In conjunction with the certification technique of
\citeauthor{cohen2019certified}, we also provide the following theorem, which  describes a bound based on the Lipschitz constant of a Gaussian averaged model. We refer to this bound as the $L$-bound, which demonstrates the link between Gaussian averaging and adversarial robustness.

\begin{definition}[$\bm{L}$\textbf{-bound}]\label{thm:bound} 
  Suppose $f^{smooth}$ is the convolution (average) of $f:\mathbb{R}^d \to [0,1]^{Nc}$ with a Gaussian kernel of variance $\sigma^2 I$,
\[
f^{smooth}(x)=\left(f\ast\mathcal{N}(0,\sigma^2 I)\right)(x)
\] 
Then any perturbation $\delta$ which results in a change of rank of the $k$-th component of $f^{smooth}(x)$ must have norm bounded as follows: 
\begin{equation}
	\left\|\delta\right\|_2 \geq {\sigma}(\pi/2)^{1/2}  (f^{smooth}(x)_{(k)}-f^{smooth}(x)_{(k+1)})
\end{equation}
where $f^{smooth}(x)_{(i)}$ is the $i^\text{th}$ largest value in the vector $f^{smooth}(x)\in[0,1]^{Nc}$.
\end{definition} See Appendix~\ref{app:boundproof} for proof.
This bound is equally applicable to deterministic or stochastically averaged
models. 
In stochastically averaged models $f^{smooth}(x)$ is replaced by the stochastic approximation of $\mathbb{E}_{\eta\sim\mathcal{N}(0,\sigma^2I)} \left[f(x+\eta)\right]$.

\subsection{Adversarial Attacks}

To test how robust our model is to adversarial examples, we calculate the minimum $\ell_2$ adversarial via our $L$-bound and we attack our models using the \textit{projected gradient descent (PGD)} \citep{kurakin2016adversarial, madry2017towards} and \textit{decoupled direction and norm (DDN)} \citep{rony2018decoupling} methods. These attacks are chosen because there is a specific way they can be applied to stochastically averaged models \citep{salman2019provably}. In the $\ell_2$ setting of both attacks, it is standard to take the step
\begin{equation}
	\label{eq:pgd}
	g = \alpha \frac{\nabla_{\delta_t}L\left(f(x+\delta_t),y\right)}{\left\|\nabla_{\delta_t}L\left(f(x+\delta_t),y\right)\right\|_2}
\end{equation}
in the iterative algorithm. Here, $x$ is an input example with corresponding
true class $y$; $\delta_t$ denotes the adversarial perturbation at its current
iteration; $L$ denotes the cross-entropy Loss function (or KL Divergence);
$\varepsilon$ is the maximum perturbation allowed; and $\alpha$ is the step-size. In the stochastically averaged model setting, the step is given by
\begin{equation}
	\label{eq:pgd-stoch}
	g_n = \alpha \frac{\sum\limits_{i=1}^n\nabla_{\delta_t}L\left(f(x+\delta_t+\eta_i),y\right)}{\left\|\sum\limits_{i=1}^n \nabla_{\delta_t}L\left(f(x+\delta_t+\eta_i),y\right)\right\|_2}
\end{equation}
where $\eta_1,\dots,\eta_n\overset{\text{iid}}{\sim}\mathcal{N}(0,\sigma^2I)$. For our deterministically averaged models, we implement the update \eqref{eq:pgd}. This is because our models are deterministic, meaning there is no need to sample noise at evaluation time. For stochastically averaged models \citep{cohen2019certified, salman2019provably}, we implement the update \eqref{eq:pgd-stoch}.

\begin{table}[b]
  \caption{$\ell_2$ adversarial distance metrics on ImageNet-1k}
  \label{tab:imagenet-adv}
  \centering
  \begin{tabular}{lcccccc}
    \toprule
    Model        & \multicolumn{2}{c}{$L$-bound}   & \multicolumn{2}{c}{PGD}  & \multicolumn{2}{c}{DDN} \\
     & median & mean & median & mean & median & mean \\
    \midrule
    \textsc{HeatSmoothing}  & 0.240 & 0.190 & 2.7591 & 2.6255 & 1.0664 & 1.2261     \\
    \textsc{SmoothAdv}  & 0.160 & 0.160 & 3.5643 & 3.0244 & 1.1537 & 1.2850  \\
    \textsc{RandomizedSmoothing} & 0.200 & 0.180 & 2.6787 & 2.5587 & 1.2114 & 1.3412  \\
    Undefended baseline & - & - & 1.0313& 1.2832 & 0.8573& 0.9864   \\
    \bottomrule
  \end{tabular}
\end{table}

\section{Experiments \& Results}

\subsection{Comparison to Stochastic Methods}

\begin{figure}[t!]
	\centering
    \begin{subfigure}{.49\textwidth}
        \centering
        \includegraphics[width=\linewidth]{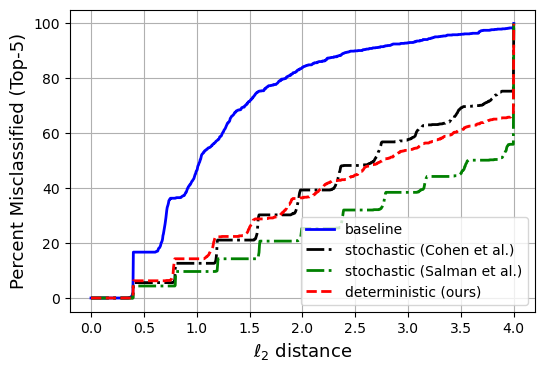}
        \caption{ImageNet-1k top-5 PGD}
    	\label{fig:imagenet-pgd}
    \end{subfigure}
    \begin{subfigure}{.49\textwidth}
        \centering
        \includegraphics[width=\linewidth]{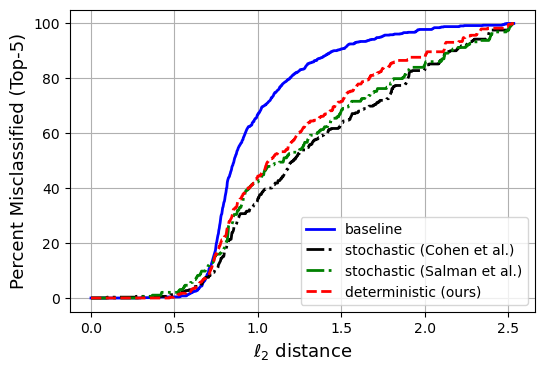}
        \caption{ImageNet-1k top-5 DDN}
    	\label{fig:imagenet-ddn}
    \end{subfigure}
    \caption{Attack curves: \% of images successfully attacked as a function of $\ell_2$ adversarial distance.}
    \label{fig:attack}
\end{figure}

We now execute our method on the ImageNet-1k dataset \citep{deng2009imagenet} with the ResNet-50 model architecture. The initial model $f$ was trained on clean images for 29 epochs with the cross-entropy loss function. Due to a lack of computing resources, we had to modify the training procedure \eqref{eq:final-variational} and Algorithm~\ref{alg:heat-smoothing} to obtain our smoothed model $f^{smooth}$. This new training procedure amounts to minimizing the loss function
\begin{equation}
	\label{imagenet-loss}
	\frac{1}{2}\left\|\textrm{softmax}\left(f^{smooth}(x+\eta)\right)-\textrm{softmax}\left(f(x)\right)\right\|_2^2 + \frac{\sigma^2}{2}\left\|\nabla_x f^{smooth}(x+\eta)\right\|_2^2
\end{equation}
for only 1 epoch of the training set using stochastic gradient descent at a fixed learning rate of 0.01 and with $\sigma=0.25$.
This is needed because the output vectors in the ImageNet setting are of length 1,000. Using softmax in the calculation of the $\ell_2$ distance metric prevents the metric from dominating the gradient-penalty term and the loss blowing up. Furthermore, we add noise $\eta\sim\mathcal{N}\left(0,\sigma^2 I\right)$ to half of the training images.

We compare our results to a pretrained \textsc{RandomizedSmoothing} ResNet-50 model with $\sigma=0.25$ provided by \cite{cohen2019certified}. We also compare to a pretrained \textsc{SmoothAdv} ResNet-50 model trained with 1 step of PGD and with a maximum perturbation of $\varepsilon=0.5$ with $\sigma=0.25$ provided by \cite{salman2019provably}. To assess certified accuracy, we run the \textsc{Certify} algorithm from \cite{cohen2019certified} with $n_0=25,n=1,000,\sigma=0.25$ for the stochastically trained models. We realize that this may not be an optimal number of noise samples, but it was the most our computational resources could handle. For the \textsc{HeatSmoothing} model, we run the same certification algorithm but without running \textsc{SamplingUnderNoise} to compute $\hat{c}_A$. For completeness, we also certify the baseline model $f^0$. Certification results on 5,000 ImageNet test images are presented in Figure~\ref{fig:imagenet-cert}. We see that our model is indeed comparable to the stochastic methods presented in earlier paper, despite the fact that we only needed one training epoch. Next, we attack our four models using PGD and DDN. For the stochastic models, we do $25$ noise samples to compute the loss. We run both attacks with max $\ell_2$ perturbation of $\epsilon=4.0$ until top-5 misclassification is achieved or until 20 steps are reached. Results on 1,000 ImageNet test images are presented in Table~\ref{tab:imagenet-adv} and Figures~\ref{fig:imagenet-pgd} and \ref{fig:imagenet-ddn}. We see that our model is comparable to the stochastic models, but does not outperform them. In Figure~\ref{fig:imagenet-pgd}, it is clear that the model presented in \cite{salman2019provably} performs best, since this model was trained on PGD-corrupted images. Note that CIFAR-10 results are presented in Appendix~\ref{app:cifar}.

%
%

\subsection{Certifying Robust Models}

So far, we have showed that we can take a non-robust baseline model and make it certifiably robust by retraining for one epoch with a regularized loss \eqref{imagenet-loss}. A natural question arises: can we use this method to make robust models certifiably robust? To test this, we begin with an adversarially trained model \citep{madry2017towards}. This pretrained model was downloaded from Madry's GitHub repository and was trained with images corrupted by the $L_2$ PGD attack with maximum perturbation size $\epsilon=3.0$. We certify this model by retraining it with \eqref{imagenet-loss} for one epoch using stochastic gradient descent with fixed learning rate 0.01. In Table~\ref{tab:cert-madry}, we compute the $\ell_2$ certified radius from \cite{cohen2019certified} for these models using 1,000 ImageNet-1k test images with $\sigma=0.25$. The certified radii for the model trained with the loss function \eqref{imagenet-loss} are significantly higher than those of the adversarially trained model from \cite{madry2017towards}.

\begin{table}[t]
  \caption{$\ell_2$ certified radii summary statistics for robust models on ImageNet-1k}
  \label{tab:cert-madry}
  \centering
  \begin{tabular}{lccc}
    \toprule
    Model        & \multicolumn{3}{c}{$\ell_2$ radius}    \\
     & median & mean & max. \\
    \midrule
    Certified adversarially trained & 0.4226 & 0.4193 & 0.6158      \\
    Adversarially trained & 0.0790 & 0.1126 & 0.6158   \\
    Undefended baseline & 0.0 & 0.1446 & 0.6158  \\
    \bottomrule
  \end{tabular}
\end{table}

\section{Conclusion}

Randomized smoothing is a well-known method to achieve a Gaussian average of
some initial neural network. This is desirable to guarantee that a model's
predictions are unchanged given perturbed input data. In this work, we used a
regularized loss to obtain deterministic Gaussian averaged models. By computing
$\ell_2$ certified radii, we showed that our method is comparable to
previously-known stochastic methods. This is confirmed by attacking our models, which results in adversarial distances similar to those seen with stochastically smoothed models. We also developed a new lower bound
on perturbations necessary to throw off averaged models, and used it as a
measure of model robustness. Lastly, our method is less computationally
expensive in terms of inference time (see Table~\ref{tab:comp-time}).
%

%


\newpage
\small
\bibliography{bibliography}
\bibliographystyle{iclr2021_conference}

\normalsize
\appendix

\newpage


\section{Solving the heat equation by training with a regularized loss function}
\label{app:proof-heat-thm}
Theorem~\ref{thm:heat-equiv} tells us that training a model with added Gaussian noise is equivalent to training a model to solve the heat equation. We can discretize the heat equation \eqref{eq:heat-eq} to obtain
\begin{equation}
    \label{eq:discreteheat}
    \frac{f^{k+1}-f^k}{h}=\frac{\sigma^2}{2}\Delta f^{k+1} 
\end{equation}
for $k=0,\dots,n_T-1$, where $n_T$ is the fixed number of timesteps, $h={1}/{n_T}$, and $f^0=f$, our initial model. Notice how, using the Euler-Lagrange equation, we can express $f^{k+1}$ in \eqref{eq:discreteheat} as the variational problem
\begin{equation}
    \label{eq:variational1}
    f^{k+1} = \underset{v}{\mathrm{argmin}} \;\frac{1}{2}\int\limits_{\mathbb{R}^d} \left(\left|v(x)-f^k(x)\right|^2 + \frac{h\sigma^2}{2}\left\|\nabla_x v(x)\right\|_2^2\right)\rho(x) dx 
\end{equation}
where $\rho$ is the density from which our clean data comes form. Therefore, this is equivalent to solving
\begin{equation}
    \label{eq:variational2}
    f^{k+1}=\underset{v}{\mathrm{argmin}} \;\mathbb{E}_x\left[ \left|v(x)-f^k(x)\right|^2 + \frac{h\sigma^2}{2}\left\|\nabla_x v(x)\right\|_2^2\right]
\end{equation}
Note that the minimizer of the objective of \eqref{eq:variational1} satisfies 
\begin{equation}
    \label{eq:min-variational}
    v-f^k = \frac{h\sigma^2}{2}\Delta v
\end{equation}
which matches \eqref{eq:discreteheat} if we set $f^{k+1}=v$. In the derivation of \eqref{eq:min-variational}, we take for granted the fact that empirically, $\rho$ is approximately uniform and is therefore constant. In the end, we iteratively compute \eqref{eq:variational2} and obtain models $f^1,\dots,f^{n_T}$, setting $v=f^{n_T}$, our smoothed model.

Something to take note of is that our model outputs be vectors whose length corresponds to the total number of classes; therefore, the objective function in \eqref{eq:variational2} will not be suitable for vector-valued outputs $f^k(x)$ and $v(x)$. We instead use the following update
\begin{equation}
    \label{eq:vector-variational}
    f^{k+1}=\underset{v}{\mathrm{argmin}} \;\mathbb{E}_x\left[ \frac{1}{2}\left\|v(x)-f^k(x)\right\|_2^2 + \frac{h\sigma^2}{2}\left\|\nabla_x v(x)\right\|_2^2\right]
\end{equation}

\section{Approximating the gradient-norm regularization term}
\label{app:jl}

By the Johnson-Lindenstrauss Lemma \citep{johnson1984extensions,vempala2005johnson-lindenstrauss}, $\left\|\nabla_x v(x)\right\|_2^2$ has the following approximation,
\begin{align}
	\begin{split}
        \left\|\nabla_x v(x)\right\|_2^2&\approx\sum\limits_{i=1}^{\kappa} \left\|\nabla_x \left(w_i\cdot v(x)\right)\right\|^2_2\\
        &\approx \sum\limits_{i=1}^{\kappa}\left(\frac{\left(w_i \cdot v\left(x+\delta \hat{g}_i\right)\right)-\left(w_i\cdot v(x)\right)}{\delta}\right)^2
       \end{split}
\end{align}
where 
\begin{equation}
    \label{eq:w}
    w_i = \frac{1}{\sqrt{K}}\left(w_{i1},\dots,w_{iK}\right)^T\in\mathbb{R}^K\;,\;\;w_{ij}\overset{\text{iid}}{\sim}\mathcal{N}(0,1)
\end{equation}
and $l$ is given by
\begin{equation}
	\label{eq:hatg}
    \hat{g}_i = \begin{cases}\frac{\nabla_x \left(w_i\cdot v(x)\right)}{\left\|\nabla_x \left(w_i\cdot v(x)\right)\right\|_2}&\text{if}\;\;\nabla_x \left(w_i\cdot v(x)\right)\neq 0\\0&\text{otherwise}\end{cases}
\end{equation}
In practice, we set $\delta=0.1$, $\kappa=10$, and $K=Nc$, the total number of classes. 

\section{Proof of Theorem~\ref{thm:bound}}
\label{app:boundproof}
\begin{proof}
Suppose the loss function $\ell$ is Lipschitz continuous with respect to model input $x$, with
Lipschitz constant $L$. Let $\ell_0$ be such that if $\ell(x)<\ell_0$, the model is always correct. Then by Proposition 2.2 in \cite{finlay2019scaleable}, a lower bound on the minimum magnitude of perturbation $\delta$ necessary to adversarially perturb an image $x$ is given by
\begin{equation}
	\label{eq:finlaybound}
	\left\|\delta\right\|_2\geq\frac{\max\left\{\ell_0 - \ell(x),0\right\}}{L}
\end{equation}
By Lemma 1 of Appendix A in \cite{salman2019provably}, our averaged model $$f^{smooth}(x)=\left(f\ast\mathcal{N}(0,\sigma^2I)\right)(x)$$ has Lipschitz constant $L=\frac{1}{\sigma}\sqrt{\frac{2}{\pi}}$. Replacing $L$ in \eqref{eq:finlaybound} with $\frac{1}{\sigma}\sqrt{\frac{2}{\pi}}$ and setting $\ell_0 = f^{smooth}(x)_{(k)}$, $\ell(x) = f^{smooth}(x)_{(k+1)}$ gives us the proof. \end{proof}

\section{Illustration of regularized training}
\label{app:train-plot}

\begin{figure}[h]
	\centering
  \includegraphics[width=.5\linewidth]{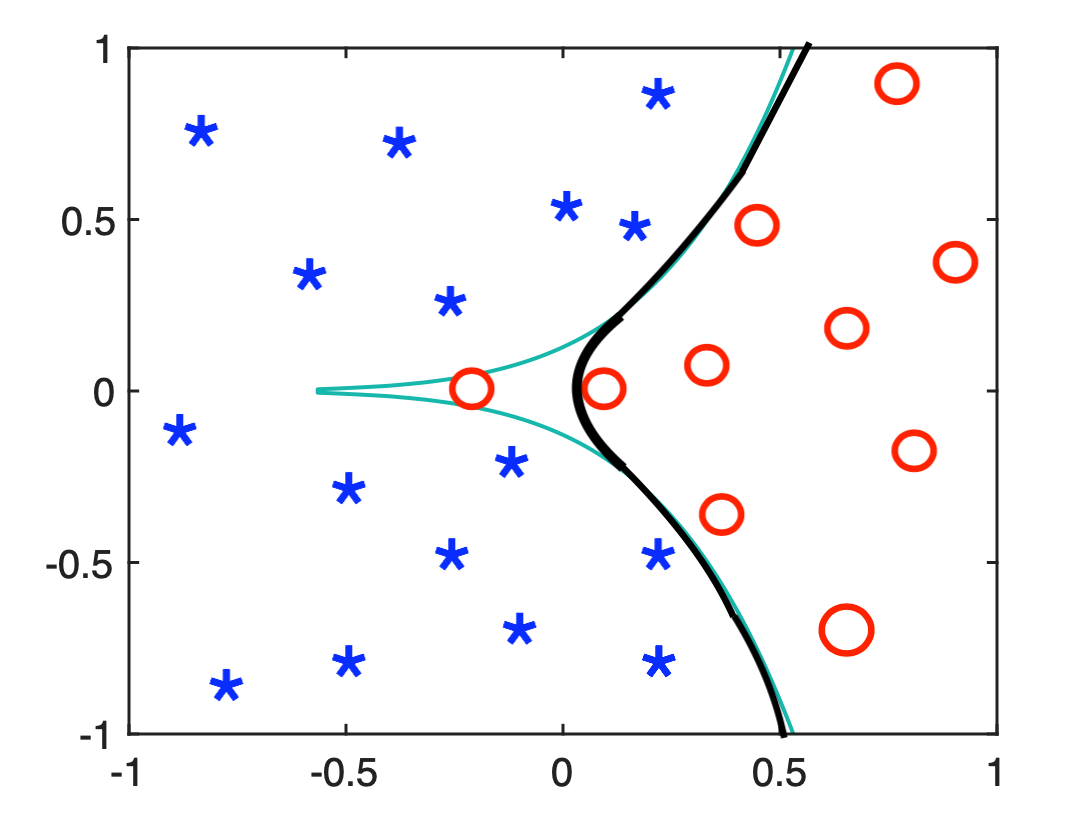} 
         \caption{Illustration of gradient regularization \eqref{Noise_Tychonoff} in the binary classification setting. The lighter line represents classification boundary for original model with large gradients, and the darker line represents classification boundary of the smoothed model.  The symbols indicate the classification by the original model: a single red circle is very close to many blue stars.  The smoothed model has a smoother classification boundary which flips the classification of the outlier.  }
        \label{fig:Cartoon}   
\end{figure}

\begin{figure}[h]
	\centering
  	\begin{subfigure}{.33\textwidth}
        \centering
        \includegraphics[width=\linewidth]{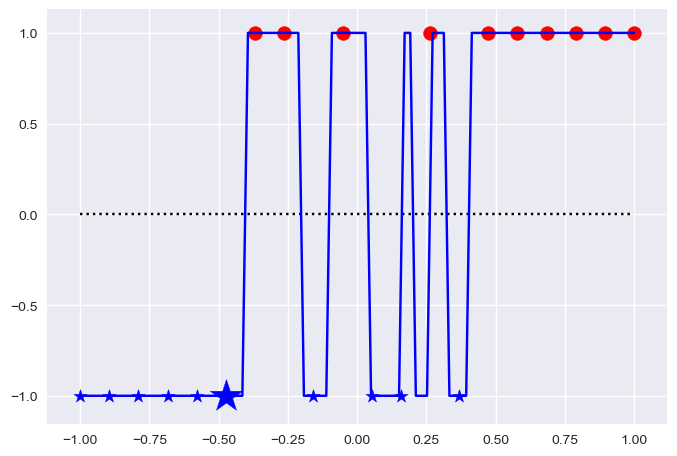}
        \caption{$f^0$}
        \label{fig:f0}
    \end{subfigure}%
    \begin{subfigure}{.32\textwidth}
        \centering
        \includegraphics[width=\linewidth]{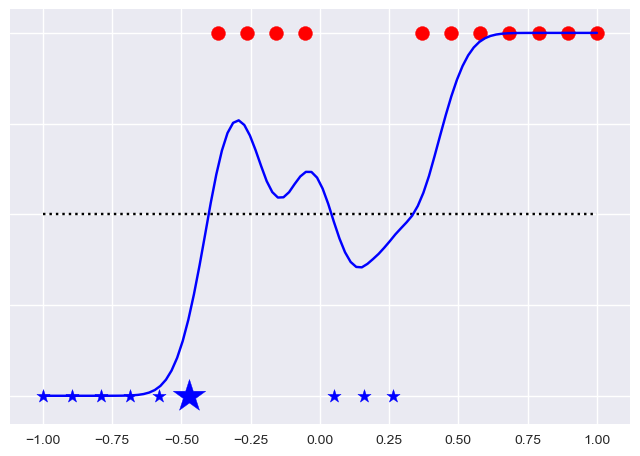}
        \caption{$f^1$}
        \label{fig:f1}
    \end{subfigure}
    \begin{subfigure}{.32\textwidth}
        \centering
        \includegraphics[width=\linewidth]{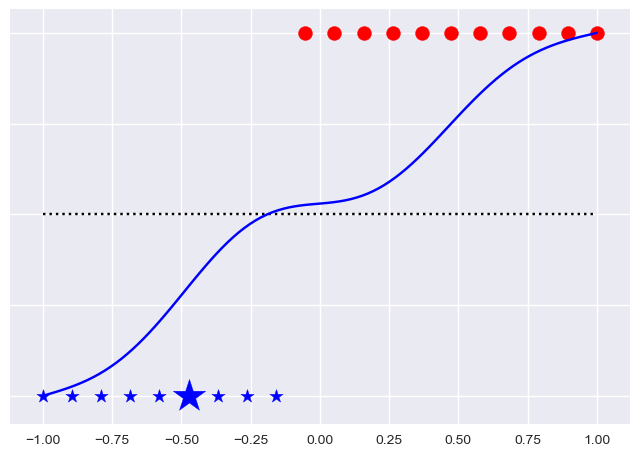}
        \caption{$f^8$}
        \label{fig:f1}
    \end{subfigure}
    \caption{Illustration of performing the iterative model update \eqref{eq:vector-variational} for 8 timesteps in the binary classification setting. The dashed black line represents our decision boundary.   The blue line represents our current classification model. The blue stars and red circles represent our predicted classes using the current model iteration. Consider the datapoint at $x=-0.5$.  In the initial model $f^0$, the adversarial distance is $\approx0.10$. In model $f^5$, the adversarial distance is increased to $\approx0.35$.}
    \label{fig:heat-eqn-solv}   
\end{figure}

\section{Results on the CIFAR-10 dataset}
\label{app:cifar}

\begin{figure}[H]
	\centering
    \begin{subfigure}{.49\textwidth}
        \centering
        \includegraphics[width=\linewidth]{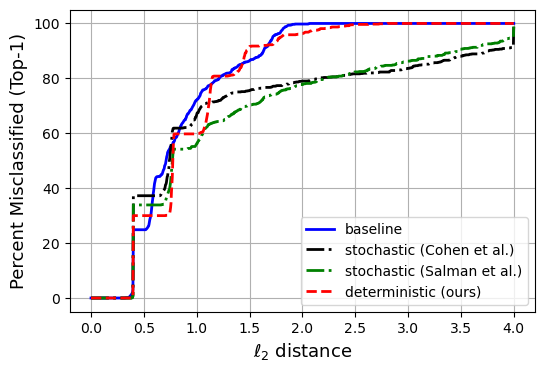}
        \caption{CIFAR-10 top-1 PGD}
        \label{fig:cifar-pgd}
    \end{subfigure}
    \begin{subfigure}{.49\textwidth}
        \centering
        \includegraphics[width=\linewidth]{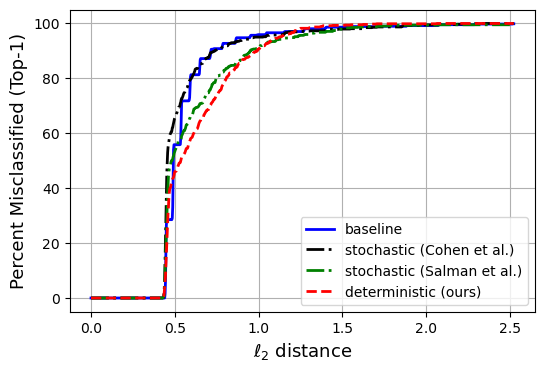}
        \caption{CIFAR-10 top-1 DDN}
    	\label{fig:cifar-ddn}
    \end{subfigure}
    \caption{CIFAR-10 attack curves: \% of images successfully attacked as a function of $\ell_2$ adversarial distance.}
    \label{fig:cifarattack}
\end{figure}

We test our method on the CIFAR-10 dataset \citep{krizhevsky2009cifar10} with the ResNet-34 model architecture. The initial model $f$ was trained for 200 epochs with the cross-entropy loss function. Our smoothed model $v$ was computed by setting $f^0=f$ and running Algorithm~\ref{alg:heat-smoothing} to minimize the loss \eqref{eq:vector-variational} with $\sigma=0.1$ for $n_T=5$ timesteps at 200 epochs each timestep. The training of our smoothed model took 5 times longer than the baseline model. We compare our results to a ResNet-34 model trained with $\sigma=0.1$ noisy examples as stochastically averaged model using \textsc{RandomizedSmoothing} \citep{cohen2019certified}. We also trained a \textsc{SmoothAdv} model \citep{salman2019provably} for 4 steps of PGD with the maximum perturbation set to $\varepsilon=0.5$. To assess certified accuracy, we run the \textsc{Certify} algorithm from \cite{cohen2019certified} with $n_0=100,n=10,000,\sigma=0.1$ for the stochastically trained models. For the \textsc{HeatSmoothing} model, we run the same certification algorithm, but without running \textsc{SamplingUnderNoise} to compute $\hat{c}_A$. For completeness, we also certify the baseline model $f^0$. Certification plots are presented in Figure~\ref{fig:cifar-cert}. In this plot, we see that our model's $\ell_2$ certified accuracy outperforms the stochastic models. Next, we attack our four models using PGD and DDN. For the stochastic models, we do $100$ noise samples to compute the loss. We run both attacks with 20 steps and maximum perturbation $\varepsilon=4.0$ to force top-1 misclassification. Results are presented in Table~\ref{tab:cifar10-adv} and Figures~\ref{fig:cifar-pgd} and \ref{fig:cifar-ddn}. In Table~\ref{tab:cifar10-adv}, we see that \textsc{HeatSmoothing} outperforms the stochastic models in terms of robustness. The only exception is robustness to mean PGD perturbations. This is shown in Figures~\ref{fig:cifar-pgd}. Our model performs well up to an $\ell_2$ PGD perturbation of just above 1.0. 

\begin{table}[H]
  \caption{$\ell_2$ adversarial distance metrics on CIFAR-10. A larger distance implies a more robust model.}
  \label{tab:cifar10-adv}
  \centering
  \begin{tabular}{lcccccc}
    \toprule
    Model        & \multicolumn{2}{c}{$L$-bound}   & \multicolumn{2}{c}{PGD}  & \multicolumn{2}{c}{DDN} \\
     & median & mean & median & mean & median & mean  \\
    \midrule
    \textsc{HeatSmoothing} & 0.094 & 0.085  & 0.7736 &0.9023 & 0.5358 &0.6361    \\
    \textsc{SmoothAdv}  & 0.090 & 0.078 & 0.7697 & 1.3241 & 0.4812 & 0.6208  \\
    \textsc{RandomizedSmoothing}  & 0.087 & 0.081 & 0.7425 &1.2677 & 0.4546 &0.5558  \\
    Undefended baseline & - & - & 0.7088 & 0.8390 & 0.4911 &0.5713 \\
    \bottomrule
  \end{tabular}
\end{table}

\end{document}